# IndicGEC: Powerful Models, or a Measurement Mirage?

**Sowmya Vajjala**
National Research Council, Canada
sowmya.vajjala@nrc-cnrc.gc.ca

## Abstract

In this paper, we report the results of the TeamNRC's participation in the BHASHA-Task 1 Grammatical Error Correction shared task[1] for 5 Indian languages. Our approach, focusing on zero/few-shot prompting of language models of varying sizes (4B to large proprietary models) achieved a Rank 4 in Telugu and Rank 2 in Hindi with GLEU scores of 83.78 and 84.31 respectively. In this paper, we extend the experiments to the other three languages of the shared task - Tamil, Malayalam and Bangla, and take a closer look at the data quality and evaluation metric used. Our results primarily highlight the potential of small language models, and summarize the concerns related to creating good quality datasets and appropriate metrics for this task that are suitable for Indian language scripts.

## 1 Introduction

Grammatical Error Correction systems aim to identify and correct the grammatical errors in a given text. This has been a task of interest in the NLP community for over a decade now (e.g., (Ng et al., 2014; Qorib and Ng, 2022). Although research on this topic is heavily dominated by English, there is some recent interest in expanding to other languages and the recent MultiGEC (Masciolini et al., 2025) shared task released datasets for GEC in 12 European languages. In comparison, Indian languages are under-represented in the research on GEC in NLP, and the Indic GEC shared task promises to fill that void in terms of resource creation and benchmarking.

**BHASHA Indic GEC Shared Task:** The BHASHA IndicGEC shared task (Bhattacharyya and Bhattacharya, 2025) aims to benchmark grammatical error detection and correction systems for Indian languages, and covers five languages: Hindi, Telugu, Tamil, Malayalam and Bangla. The goal of the task is to correct grammatical errors (if any) in the input sentence and generate the output sentence. Participants were provided varying amounts of training/development sets across languages and the test sets released later were used to compare and rank the systems. GLEU score (Napoles et al., 2015), which is based on n-gram overlap between the candidate sentence and the reference sentence, was used for evaluating and comparing across the systems.

**Our Approach and Contributions:** Our approach is primarily focused on understanding the capabilities of LLMs of various sizes in zero-shot and few-shot settings when prompted to reason and act. In the final leader board, this approach stood 4th for Telugu (zero-shot with Gemma3:12B model) and 2nd for Hindi (zero-shot with Gemini-2.5-Flash model). In this system description paper,

1. we report on the primary approach followed and the results for Telugu and Hindi, where we participated,

2. describe the results of the extended experiments covering the other three languages of the shared task,

3. highlight the existing data quality issues in Telugu, and

4. argue for a better evaluation metric for this task by qualitative analysis.

Apart from providing a window into the capabilities and limitations of off-the-shelf

---

[1] https://github.com/BHASHA-Workshop/IndicGEC2025/

LLMs for Indian language GEC, We hope that these observations lead to improving the resource creation and evaluation aspects of GEC for Indian languages in the future.

Section 2 describes the overall approach used including the choice of LLMs, prompting and choosing the few-shot samples. Section 4 contains the quantitative results across languages, and Section 4 addresses the data quality issues taking Telugu as the test case, and also discusses the suitability of GLEU as the metric for this task/these languages. Section 5 summarizes the main results and challenges with doing GEC for Indian languages.

## 2 Approach

Our objective is to explore the off-the-shelf in-context learning capabilities of existing multilingual LLMs of varying sizes without resorting to any task-specific finetuning. We believe this is an important (and as of now missing) step to evaluate the abilities of current LLMs for performing grammatical error correction in Indian languages. To this end, we experimented with the Gemma3 (Team et al., 2025) models of three sizes - 4B, 12B, 27B, in a zero-shot setting. To compare with a larger model, we used gemini-2.5-flash (Comanici et al., 2025) in the same setting. For the small multilingual models that supported all or most of the Indian languages in the shared task, we initially compared Qwen3 (Yang et al., 2025) series of models and Sarvam-M[2], and chose the Gemma3 models as the inference was faster and consistent (without throwing server access errors). Both the smaller models (4B and 12B) were also evaluated in a k-shot setting (k=0,5,10,15) where the examples were randomly selected from the provided language specific training data. We did not consider few-shot learning for larger models due to time constraints.

All the models were accessed through OpenRouter[3] in a ReAct (Yao et al.) setup, which essentially prompts an LLM to generate reasoning traces that help the model to perform the given task. The same experimental setup was repeated for the five languages separately, and we did not explore any cross-lingual transfer in this paper. Full implementation code is shared in Appendix, in Figure 2 for one language/LLM, and remained the same for all languages except for swapping the language names in the function descriptions. The provided training data was only used to collect the randomly chosen few-shot samples and no further task-specific fine-tuning was performed for any language. The total experimental costs amounted to 10 USD.

## 3 Results

The performance of all the models in the settings described in Section 2 was evaluated using the official GLEU metric script provided, and the results are discussed in this section. Figure 1 shows the results across all the experimental settings for all the five languages - Telugu, Hindi, Tamil, Malayalam, Bangla. Note that the test sets for individual languages are not evenly distributed, with Telugu and Bangla test-sets having over 300 sentences, Hindi and Malayalam between 150-250 sentences, and Tamil with only 65 sentences.

While we can notice performance difference among the different models across languages in zero-shot settings, the differences are less dramatic than expected considering the large variation in the model parameter sizes. That indicates that even the very small 4B model has some notion of grammar in various Indian languages, and reaches a GLEU score of almost 90 for Malayalam. Gemini-2.5-Flash, the largest model among the four, gave the highest zero-shot performance for all languages except Telugu, where the 12B model scored better. Few-shot learning seemed to be somewhat useful for Malayalam and Bangla, but otherwise seemed to harm the models than help them do better on the task.

We did not explore efficient sampling techniques for few-shot sampling, and took random k-samples from the training set. A more thoughtful choice of the training examples (e.g., choosing the ones with more character transformations instead of the ones with one or two minor changes) may help in boosting the few-shot performance. Few-shot sampling for Gemma3-12B model in one case (k=15 for Bangla) went past the zero-shot performance of the much larger gemini-2.5-flash model by over 1.5%, which indicates exploring better sampling strategies may be useful.

---

[2]sarvam.ai
[3]https://openrouter.ai/

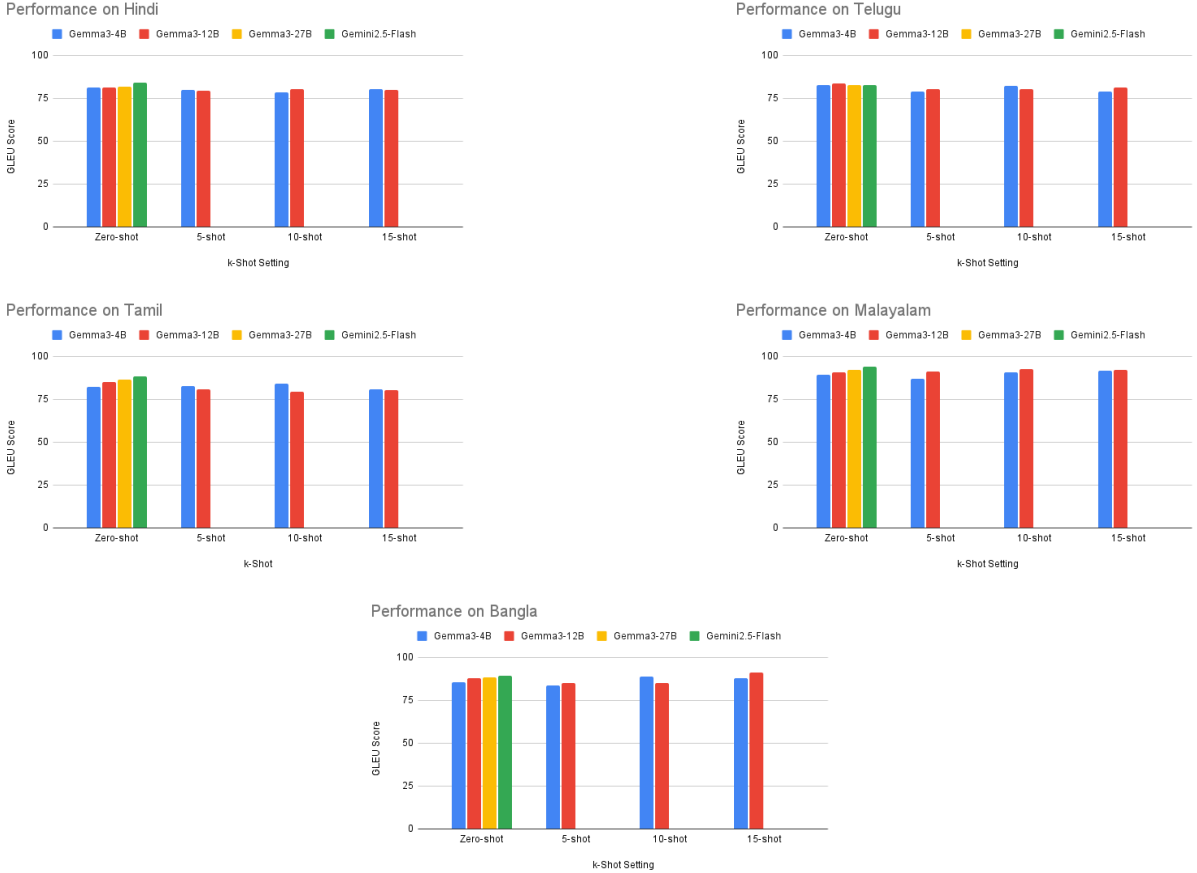

Figure 1: k-shot performance of 4 LLMs across the 5 languages

## 4 Qualitative Analysis

To understand why few-shot prompting did not help and resulted in lower performance in some cases, we did some qualitative analysis choosing Telugu as the target data analysis language. The goal of this analysis is to understand the nature of the transformations between input and expected output. This section summarizes some of the observations on the data quality in the Telugu subset of the shared task's datasets.

We analyzed a sample of 100 test sentences, and about 30% of the sentences had some error in the gold output. We identified four broad issues from this analysis, illustrated below with one example for each category:

**A. Wrong Outputs:** There were several examples where the gold standard output had either spelling or grammatical errors. As an example, consider this input sentence and gold output pair: దాంట్లో పాండవులు విజిం దకుతుంది. −> దాంట్లో పాండవులకి విజఅం దక్కుతుంది.. The gold output has a spelling error and the corrected output is: దాంట్లో పాండవులకి <u>విజయం</u> దక్కుతుంది..

**B. Sentences that need additional context:** In some cases, the context for the input (e.g., the correct tense, the right usage of honorific, verbal inflection for gender etc) came from the previous sentence in the test set. However, the task itself assumed no such context, as the inputs were individual sentences. For example, consider this input-gold output pair: అందరితో బాగా కలుస్తారు. −> అందరితో బాగా కలుస్తాడు.. In this pair, both are grammatically correct. We cannot choose one over the other without knowing the previous context on whether the person referred to in the verb is being referred to as "they" (as in input) or "he" (as in the gold output).

**C. Formatting issues:** There were also a few formatting issues such as insertion of extraneous characters in the output (or input), and having incomplete sentences. For example, consider this input-gold output pair: ఈమధ్యనే "దొమ్మెట్టి గుకేష్.–> ఈమధ్యనే "దొమ్మెట్టి గుకేష్. This translates to: *Recently, "Dommetti Gukesh..* Firstly, this is an incomplete sentence. Secondly, this is potentially referring to the well-known Indian chess player Dommaraju Gukesh, and there is an error in the proper name, where దొమ్మెట్టి should be దొమ్మరాజు. Whether spelling errors in proper names come under GEC or not is a separate question to address, but there were several examples of this kind in the training set as well.

**D. Multiple Correct Outputs:** Finally, there were also many examples where multiple correct outputs are possible. For example, consider the following input-gold output pair: పరశురాముడు శ్రీ మహా విష్ణువు యొక్క ఆరవ అవతారము. –> పరశురాముడు శ్రీమహావిష్ణువు యొక్క ఆరవ అవతారం.. Both the sentence endings are grammatically correct in Telugu, and having a space in మహా విష్ణువు is not an error in itself. Thus, the input is not a gramamatically incorrect sentence. Note that this classification of errors is only based on the analysis of 100 sentences, and a more fine-grained definition of errors may be necessary in future to understand the model performance better.

### 4.1 Reasoning traces of the models may help

Interestingly, looking through the models' reasoning traces for such examples reveal that the models are capable of identifying such issues in the datasets. For example, for the sentence with formatting issues above, the Gemma3-12B model's reasoning trace reads as *The input sentence ఈమధ్యనే "దొమ్మెట్టి గుకేష్. is incomplete and lacks a proper ending. It appears to be a fragment of a sentence. Without further context or the intended completion, it's impossible to correct it grammatically. Therefore, I will return the original input as is..* For the examples in 2 and 4, the model identifies them as grammatically correct. Such reasoning traces from LLMs could be a potential source of identifying quality issues in the GEC datasets in a semi-automatic manner, which we leave as an exploration for the future.

Note that we did not do a full manual analysis of the entire Telugu subset, as that is beyond the scope of this paper. But this brief analysis revealed multiple errors in the test set, and the possibility of using LLM reasoning traces to identify such issues. This analysis leads us to conclude that more rigorous data quality assurance methods need to be incorporated early on into this process, to make the datasets and conclusions based on them more meaningful.

### 4.2 Suitability of GLEU as a Metric

In the subsequent analysis, we looked at the suitability of the GLEU metric for this task. GLEU score is based on n-gram matches between character sequences, which may be unsuitable considering the alpha-syllabic nature of the Indian language scripts. Hence, We did a brief analysis considering minor variations of a single Telugu sentence, all with one or two unicode character errors from the input sentence. Table 1 shows the seven manually created variations of the gold output and their respective GLEU scores with the gold output, using the official scoring script. Although all these errors can be perceived as equivalent to each other in terms of the corrections required, the metric's scores vary substantially across sentences.

This is not an issue specific to Telugu, though. Table 2 shows the GLEU score when the input is returned as-is without any changes or without using any model in between, which we can consider as a random baseline, across all languages. If we compare this with our results across models in Section 3, it is clear that all the models perform well-below this random baseline for all languages except Hindi. For Tamil, Bangla and Malalayam, just returning the input without doing anything else gives a GLEU score close to 95, which raises questions on the usefulness of the GLEU score for this task.

Although GLEU is a commonly used metric for GEC, it is not free from criticism and several other alternative metrics have been proposed over the years. Goto et al. (2025) recently developed a library to compare across the different GEC metrics. Extending the evaluation to cover more metrics and drawing comparisons across the metrics may help in choosing a better metric for IndicGEC in future.

| Gold Output Sentence: ఇది తెలుగులో తేలికగా రాసేందుకు ఉపకరిస్తుంది |  |
|---|---|
| **Output Sentence** | **GLEU score** |
| ఇది తెలుగులో తేలికగ రాసేందుకు ఉపకరిస్తుంది | 92.18 |
| ఇది తలుగులో తేలికగా రాసేందుకు ఉపకరిస్తుంది | 86.82 |
| ఇది తెలుగులో తేలికగా రాసేందకు ఉపకరిస్తుంది | 86.01 |
| ఇది తెలుగులో తేలికగా రాసేందుకు ఉపకరస్తుంది | 86.01 |
| ఇది తెలుగు లో తేలికగా రాసేందుకు ఉపకరిస్తుంది | 100.0 |
| ఇది తెలుగులో తేలికగా రాసేందుకు ఉపకరిస్తుది | 92.99 |
| ఇది తెలుగులో తేలికగా రాసేందుకు ఉపకరిస్తది | 85.33 |

Table 1: Example to illustrate variations in GLEU score. All outputs have a variation of no more than one or two unicode characters from the gold output.

| Language | GLEU score |
|---|---|
| Bangla | 95.79 |
| Hindi | 80.75 |
| Malalayam | 94.42 |
| Tamil | 92.17 |
| Telugu | 88.17 |

Table 2: GLEU score when input is the output

## 5 Conclusions

In this paper, we reported on our experiments with zero-shot and few-shot prompting of small and large language models for the IndicGEC task covering five Indian languages - Telugu, Hindi, Tamil, Malayalam, and Bangla. Our zero-shot approach stood 2nd in Hindi and 4th in Telugu among the systems that submitted during the testing phase. Extending these results to Tamil, Malayalam, and Bangla show better results for those languages. However, qualitative analysis of the data and the metric raise several issues, which lead to us back to the original question - what do the LLMs really know about this task?, without a clear answer.

Our experiments highlight the potential of the smaller multilingual language models, but also point to the data quality issues and the potentially flawed evaluation metric. Clearly, that leads to several ideas for future work starting with the development of high quality datasets and better evaluation metrics first. The datasets also should cover a diverse variety of potential grammatical errors in the language, which may need some kind of fine-grained error annotation. Once they are in place, better prompting and potentially fine-tuning, including cross-lingual transfer can be explored for this task. We hope that these experiments serve as a reality check and lead towards the development of better methodology for GEC, especially in the Indian language context, in future. As a best practice recommendation for the future, we suggest shared tasks should perhaps consider a "data vibe check" as the first phase of the competition before letting the participants compete on the models themselves.

## Limitations

Our experiments are limited to a small selection of language models, prompted in a limited setting, and the results have to be understood within these constraints.

## Author Note

This paper was originally submitted to the shared task as a system description paper, but was evaluated as out of scope for a shared task paper 1) as it has more discussion on the choice of dataset and metric, and 2) it only focuses only on zero/few-shot approaches without resorting to more advanced approaches. Nevertheless, the results offer some insights to the research community on grammatical error correction task especially for languages with non-Roman scripts.

# A  Implementation Code

```python
import dspy
lm = dspy.LM(
    model="openrouter/google/gemma-3-12b-it",
    api_base="https://openrouter.ai/api/v1",
    api_key="Your_API_KEY"
) #Or use local Ollama models without a key.
dspy.configure(lm=lm)

#Class Signature
class gechin(dspy.Signature):
    """The task is to do grammatical error correction for Hindi.
    Input is a Hindi sentence with some or no errors,
    and output should be the Hindi sentence with errors corrected.
    If there are no errors, return the same sentence."""
    source_text = dspy.InputField()
    target_text = dspy.OutputField()

#React call
class GECHin_React(dspy.Module):
    """You are a Grammatical Error Correction system for Hindi.
    Rewrite the given Hindi text correcting the grammatical
    or spelling errors in it, if there are any."""
    def __init__(self):
        super().__init__()
        self.prog = dspy.ReAct(gechin, tools=[])

    def forward(self, source):
        return self.prog(source_text=source)

#Zero-shot Prediction
zeroshot = GECHin_React()

#Example output:
output = zeroshot.forward(test_sens[0])
output_text = output.target_text
explanation = output.reasoning

#Few-shot Sampling
myfewshots = [] #will contain the few shot examples
for i in range(0,n): #n=5,10,15
    #take a random numbered training sample from a list of training sentences.
    mynum = randint(0,len(train_sens))
    myfewshots.append(dspy.Example(source=train_sens[mynum],
                output=train_outputs[mynum]).with_inputs("source"))

#Few-shot prediction
fs = GECHin_React()
bootstrap = dspy.BootstrapFewShot(max_bootstrapped_demos=len(myfewshots))
bootstrapped_student = bootstrap.compile(fs, trainset=myfewshots)

#Example output
output = bootstrapped_student(test_sens[0])
output_text = output.target_text
explanation = output.reasoning
```

Figure 2: Implementation Code